\documentclass[10pt,twocolumn,letterpaper]{article}

\usepackage{cvpr}              

\definecolor{cvprblue}{rgb}{0.21,0.49,0.74}
\usepackage[pagebackref,breaklinks,colorlinks,citecolor=cvprblue]{hyperref}

\usepackage{multirow}
\usepackage{graphicx}
\usepackage{amsmath}
\usepackage{pifont}
\usepackage{xcolor,colortbl}
\pdfoutput=1

\newcommand{\cmark}{\ding{51}}%
\newcommand{\xmark}{-}%

\newcounter{rownumbersa}

\newcounter{rownumbersb}

\def\paperID{5511} 
\def\confName{CVPR}
\def\confYear{2025}

\title{SCFlow2: Plug-and-Play Object Pose Refiner with Shape-Constraint Scene Flow}

\author{%
	{Qingyuan Wang $^1$, \quad Rui Song $^1$, \quad Jiaojiao Li $^1$, \quad Kerui Cheng $^{2}$, \quad David Ferstl $^{3}$, \quad Yinlin Hu $^{3}$} \\
	{\normalsize $^1$ State Key Laboratory of ISN, Xidian University \quad \quad $^2$ Taiyuan University of Technology \quad \quad $^3$ MagicLeap} \\
}

\begin{document}
\maketitle
\begin{abstract}
We introduce SCFlow2, a plug-and-play refinement framework for 6D object pose estimation. Most recent 6D object pose methods rely on refinement to get accurate results. However, most existing refinement methods either suffer from noises in establishing correspondences, or rely on retraining for novel objects. SCFlow2 is based on the SCFlow model designed for refinement with shape constraint, but formulates the additional depth as a regularization in the iteration via 3D scene flow for RGBD frames. The key design of SCFlow2 is an introduction of geometry constraints into the training of recurrent matching network, by combining the rigid-motion embeddings in 3D scene flow and 3D shape prior of the target. We train SCFlow2 on a combination of dataset Objaverse, GSO and ShapeNet, and evaluate on BOP datasets with novel objects. After using our method as a post-processing, most state-of-the-art methods produce significantly better results, without any retraining or fine-tuning. The source code is available at \url{https://scflow2.github.io}.
\end{abstract}

\section{Introduction}
\label{sec:intro}

6D object pose estimation, which determines an object’s 3D rotation and 3D translation relative to a camera, is crucial in fields like robotics, augmented reality, and spatial computing~\cite{marchand2015pose, wang2019densefusion}.

Most existing pose estimation methods rely on a refinement procedure to get accurate pose results~\cite{wang2019densefusion,labbe2020cosypose,labbe2022megapose,hu2022pfa,haugaard2022surfemb,hai2023pseudo,wen2024foundationpose}, and most of the refinement methods are based on render-and-compare strategies~\cite{labbe2022megapose,hu2022pfa,hai2023pseudo,wen2024foundationpose}.
A group of refinement methods~\cite{wang2019densefusion,labbe2020cosypose} achieves highly accurate results based on retraining or fine-tuning the network for novel objects, which, however, is cumbersome and unfriendly to real applications. Some recent refinement methods~\cite{labbe2022megapose,shugurov2022osop,wen2024foundationpose} have shown promising results without fine-tuning for novel objects. However, they still suffer from several weaknesses.

\begin{figure}[tp]
    \centering
    \setlength\tabcolsep{1pt}
    \begin{tabular}{c}
    \includegraphics[width=1.0\linewidth]{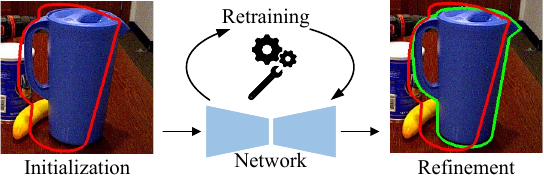} \\
    {\small (a) Object pose refinement with retraining} \\
    \includegraphics[width=1.0\linewidth]{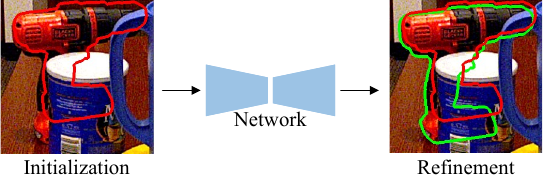} \\
    {\small (b) Plug-and-play SCFlow2} \\
    \end{tabular}
    \vspace{-1em}
    \caption{{\bf Object pose refinement} is critical for accurate object pose estimation. {\bf (a)} Most existing object pose refinement methods, including SCFlow~\cite{hai2023scflow}, rely on retraining for novel objects to achieve high accuracy. {\bf (b)} The proposed SCFlow2 achieves even higher accuracy, and more importantly, generalizes well to novel objects without any retraining or fine-tuning.
    }
    \label{fig:teaser}
\end{figure}

\begin{figure*}[tp]
    \centering
    \begin{tabular}{cc}
    \hspace{-1em}
    \includegraphics[width=0.54\linewidth]{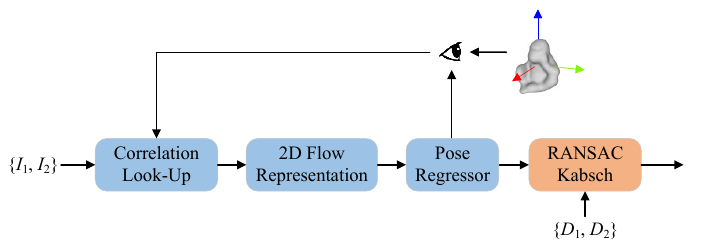} &
    \hspace{-1em}
    \includegraphics[width=0.48\linewidth]{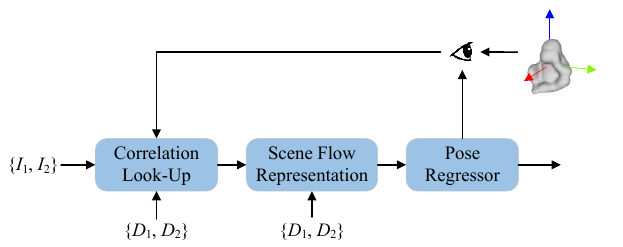} \\
    {\small (a) The framework of SCFlow} & {(b) \small The framework of SCFlow2} \\
    \end{tabular}
    
      
      
      
    \vspace{-0.5em}
    \caption{{\bf Design overview of SCFlow and SCFlow2.} Given the object 3D mesh, we render an image $I_1$ and depth map $D_1$ based on an initial pose, and then use networks to compare these rendered outputs with the real input $I_2$ and $D_2$ to refine the pose.
    {\bf (a)} Although SCFlow~\cite{hai2023scflow} adds 3D shape constraint into the optimization loop, it formulates the matching process as a pure 2D problem, which is less effective in capturing 3D motions. On the other hand, it cannot work with RGBD images. A common practice is to use RANSAC Kabsch~\cite{kabsch1976kabsch} to consume additional depth as a second stage, which however is only local optimal within each stage. {\bf (b)} SCFlow2 tackles these problems. We introduce an intermediate representation based on 3D scene flow to capture 3D motions in network optimization. Furthermore, we embed depth into the loop by formulating depth as an additional regularization to guide the correlation look-up iteratively, producing an end-to-end trainable system with RGBD images.
    }
    \label{fig:scflow_vs_scflow2_overall}
\end{figure*}

First, most methods formulate the render-and-compare framework as a general matching problem, but there is a strong prior of the target's 3D shape that is not used, making it usually has an unnecessarily large search space for matching and is less effective for 6D object pose estimation~\cite{hai2023scflow}. Second, to address the matching issue, most methods attempt to refine the pose starting from multiple pose hypotheses based on the initialization~\cite{labbe2022megapose,moon2024genflow, ornek2023foundpose}. While this paradigm is effective, it significantly slows down the refinement process. Third, most methods formulate the comparison after the rendering as a 2D matching procedure only, and then use ICP or Kabsch~\cite{kabsch1976kabsch} to consume additional depth input as a second stage, which is less effective in capturing 3D motions and local optimal within each stage.

We introduce SCFlow2, a plug-and-play refinement framework for 6D object pose estimation, as shown in Fig.~\ref{fig:teaser}. SCFlow2 is based on SCFlow~\cite{hai2023scflow} which is a state-of-the-art refinement method embedding 3D shape constraints into the optimization for known objects. We build on the basic shape-constraint structure of SCFlow but extend it in several novel ways.

Our main design of SCFlow2 is introducing a scene flow based representation into the iterative optimization. Although SCFlow embeds shape constraint into the optimization loop, it formulates the matching process as a pure 2D problem, and suffers in the solution with a second stage of RANSAC Kabsch~\cite{kabsch1976kabsch} for RGBD images.
On the other hand, SCFlow cannot work with novel objects. To handle these problems, we introduce rigid-motion embeddings in 3D scene flow~\cite{teed2021raft3d} as our intermediate representation, and formulate depth as an additional regularization to guide the network optimization iteratively, which helps reducing search space and results in an end-to-end system that generalizes to novel objects, as illustrated in Fig.~\ref{fig:scflow_vs_scflow2_overall}.

We train SCFlow2 on a combination of dataset Objaverse~\cite{deitke2023objaverse}, GSO~\cite{downs2022gso} and ShapeNet~\cite{chang2015shapenet}, and evaluate on BOP datasets~\cite{hodan2018bop} with novel objects. SCFlow2 achieves state-of-the-art accuracy, and can produce accurate results with only one pose hypothesis, outperforming most multiple-hypothesis based refinement methods~\cite{moon2024genflow,labbe2022megapose, ornek2023foundpose} in both accuracy and efficiency, without any retraining.
\section{Related Work}
\label{sec:related}

\begin{figure*}[tp]
    \centering
    \includegraphics[width=1.0\linewidth]{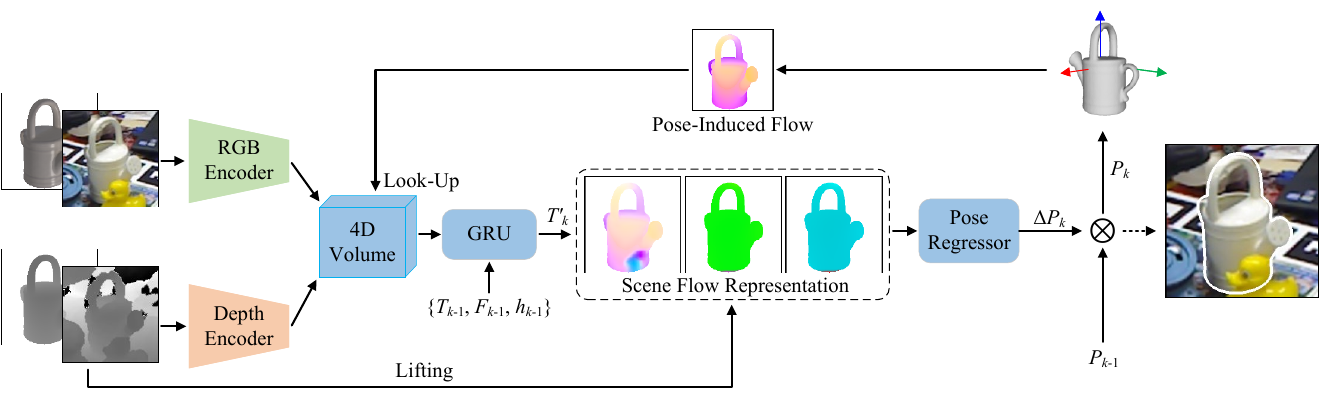}
    \caption{{\bf Overview of SCFlow2. } Given an RGBD image and an initial pose of the target, we first render the target to obtain a synthetic RGBD image as the reference, and use an RGB encoder and depth encoder to extract features from the image pair, which will be used to create a 4D correlation volume. Based on the correlation volume and GRU, We use an intermediate flow regressor to predict 3D scene flow that is represented as a dense 3D transformation field $\mathbf{T}_{k}^{\prime}$. We then use a pose regressor to predict a global pose update $\Delta\mathbf{P}_{k}$ based on implicit voting from the pixel-wise 3D transformation field. Finally, the updated pose $\mathbf{P}_{k}$ is used to compute the pose-induced flow $F_{k}$ based on the target mesh to index the correlation volume for the next iteration. Note how the depth and 3D shape of the target are embedded into the framework to guide the optimization iteratively.
    }
    \label{fig:scflow2_main}
\end{figure*}

\noindent {\bf Object pose estimation} aims to predict the 3D rotation and 3D translation with respect to the camera, and has shown prominent progress with deep learning techniques~\cite{he2022fs6d,park2019latentfusion,sun2022onepose,nguyen2024nope,lin2024sam6d, hipose}. Most existing methods assume the existence of many real images of the target, and train a learning model based on those real images~\cite{hu2021wdr, hu2022pfa, hai2023scflow, peng2019pvnet, su2022zebrapose}, based on either 2D-to-3D matching~\cite{zakharov2019dpod, li2019cdpn, hodan2020epos}, 3D-to-2D matching~\cite{rad2017bb8, hu2021wdr, peng2019pvnet}, or end-to-end training with differentiable perspective-n-point (PnP)~\cite{song2020hybridpose, chen2020bpnp, liu2023linear}. This type of method is accurate on the trained target in general, which however, cannot work with novel objects that were not involved in the training. Retraining or fine-tuning can adapt the network to novel objects, which, however, is cumbersome and limits its applications in real scenarios. Some recent methods show promising results for novel object pose estimation~\cite{labbe2022megapose, nguyen2024gigapose, moon2024genflow,ornek2023foundpose, wen2024foundationpose}, and one of their key components is object pose refinement, which is based on render-and-compare~\cite{labbe2022megapose,moon2024genflow,wen2024foundationpose} or feature matching~\cite{ornek2023foundpose, nguyen2024gigapose} strategies. Although these refinement methods work in general, they are less effective at utilizing prior shape information of the target. The recent method SCFlow~\cite{hai2023scflow} introduces a shape-constraint recurrent matching framework for object pose estimation, which embeds the target's 3D shape into the optimization. It produces accurate results for known objects, while has limited generalization ability to novel objects. SCFlow2 is built on top of the basic shape-constraint structure of SCFlow, but extends it by introducing 3D scene flow representation into the network optimization, producing an end-to-end system with RGBD images.

\noindent {\bf Optical flow estimation} is a fundamental computer vision task to estimate pixel-level correspondences between two frames~\cite{horn1981determining,saxena2024surprising,cuadrado2023optical,2024LowlightF}. Most recent methods are based on deep learning models and achieve accurate results in general scenes~\cite{teed2020raft,cuadrado2023optical,saxena2024surprising}. The recent method RAFT~\cite{teed2020raft} presents a recurrent framework for flow estimation, and RAFT-3D~\cite{teed2021raft3d} extends it to estimate 3D scene flow with rigid motion embeddings. Although they show significant advantages over existing flow methods, estimating dense matching with these general flow methods for object pose estimation is less effective, without using the prior 3D shape information of the target. Furthermore, the general flow loss is a surrogate matching loss that does not directly reflect the final 6D object pose~\cite{hu2022pfa,hai2023scflow,hu2020single,song2020hybridpose} in the context of object pose estimation. We combine the shape-constraint formulation~\cite{hai2023scflow} and the rigid motion embeddings in 3D scene flow, producing a geometry-guided 6D object pose refiner that is end-to-end trainable and generalizes to novel objects.

\section{Method}
\label{sec:method}


Given an RGBD image and the 3D mesh of the object, our goal is to estimate the 6D pose of the target object. We assume the camera intrinsic matrix is known. We first introduce the framework at a high level in \Cref{subsec:overview}, and then present the design of the main components in \Cref{subsec:corrV,subsec:se3motiton}. We finally discuss the implementation details in \Cref{subsec:details}.




\subsection{Overview}
\label{subsec:overview}

In general, we first render an RGBD image based on an initial pose of the target as a reference, and then predict the pose update based on comparing the rendered image and the real input with a recurrent matching network.

With the RGBD image pair, we first use an RGB encoder and depth encoder to extract features, and then construct a 4D correlation volume based on computing the dot product of the fused RGBD features. Based on the constructed correlation volume, we propose to use an intermediate flow regressor to predict 3D scene flow that is represented as a dense SE3 motion field~\cite{teed2021raft3d}. We then use a pose regressor to predict a global pose update based on implicit voting from the pixel-wise SE3 motion field. Finally, the updated pose is used to compute a pose-induced flow based on the target mesh to index the correlation volume for the next iteration.

The key of our method is the combination of the dense 3D scene flow representation and the embedding of the target 3D shape prior~\cite{hai2023scflow} in a unified recurrent matching framework. Fig.~\ref{fig:scflow2_main} shows the overview of this framework.



\begin{figure*}[tp]
    \centering
    \setlength\tabcolsep{1pt}
    \begin{tabular}{ccccccc}


    \includegraphics[width=0.14\linewidth]
    {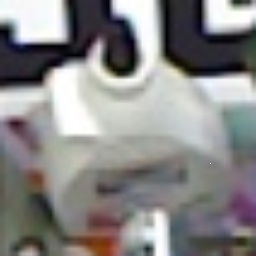} &
    \includegraphics[width=0.14\linewidth]
    {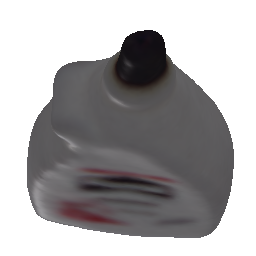} &
    \includegraphics[width=0.14\linewidth]
    {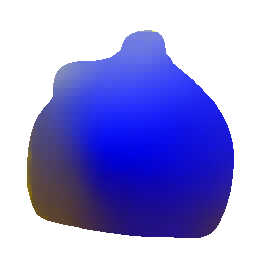} &
    \includegraphics[width=0.14\linewidth]
    {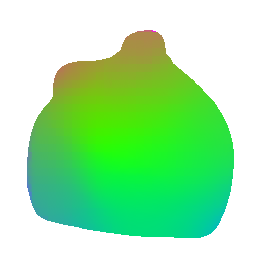} &
    \includegraphics[width=0.14\linewidth]
    {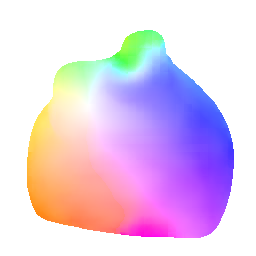} &
    \includegraphics[width=0.14\linewidth]
    {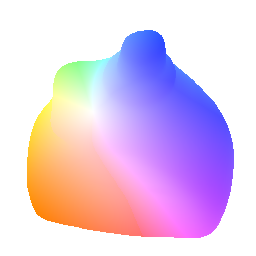} &
    \includegraphics[width=0.14\linewidth]
    {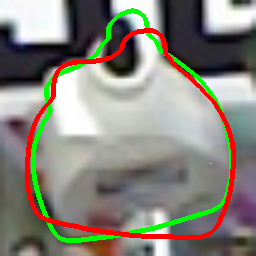} 
    \\

    \includegraphics[width=0.14\linewidth]
    {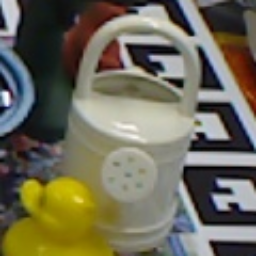} &
    \includegraphics[width=0.14\linewidth]
    {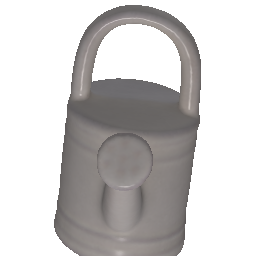} &
    \includegraphics[width=0.14\linewidth]
    {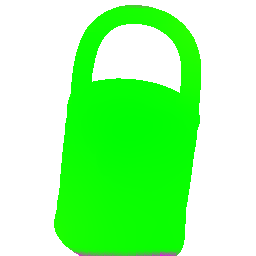} &
    \includegraphics[width=0.14\linewidth]
    {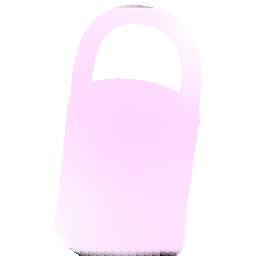} &
        \includegraphics[width=0.14\linewidth]
    {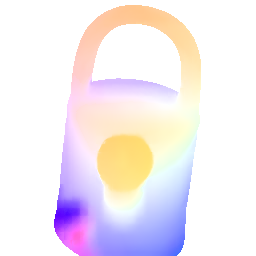} &
    \includegraphics[width=0.14\linewidth]
    {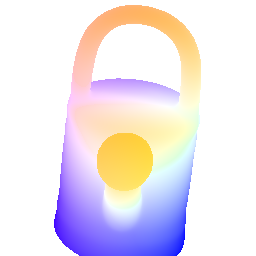} &
    \includegraphics[width=0.14\linewidth]
    {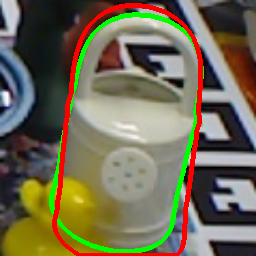} 
    \\

    {\small Input} & {\small Initial pose} & {\small $\tau$} & {\small $\phi$} & {\small Intermediate flow} & {\small Pose-induced flow} & {\small Refined pose}\\
    \end{tabular}
    \caption{{\bf Visualization of pose-induced flow with 3D scene flow representation. } Given a real input image and the rendered image based on an initial pose, we predict 3D scene flow represented as a dense SE3 motion field. We represent the motion field as a twist field ($\tau$, $\theta$). In theory, the motion fields should be constant pixels for rigid objects. We use a pose regressor to predict a global object-level pose based on the noisy motion fields. We then use the updated global pose to generate a pose-induced flow based on the target 3D mesh. The pose-induced flow embeds the shape prior of the target and reduces the search space for matching.
 }
    \label{fig:scene_flow}
\end{figure*}

\subsection{Correlation Volume Construction}
\label{subsec:corrV}

We construct a 4D volume pyramid to represent the correlation between any two locations on the two input images. We use an RGB encoder and a depth encoder to extract features from the RGB and the depth image, respectively. For the rendered image $I_1$ based on the initial pose and the input RGB image $I_2$, we use an encoder with shared parameters to obtain their dense feature maps $f_{1}^{i}\in\mathbb{R}^{H\times W\times C}$ and $f_{2}^{i}\in\mathbb{R}^{H\times W\times C}$ at $1/8$ of the original resolution. We use DINOv2 ViT-B~\cite{oquab2023dinov2} as our RGB feature extractor in the encoder, and freeze the pretrained weights during training. For the depths $D_1$ and $D_2$, we first lift the two depth maps to point clouds according to the camera intrinsic matrix, and then extract their 3D structural features with PointNet++~\cite{qi2017pointnet++}.

Similar to DenseFusion~\cite{wang2019densefusion}, we fuse the RGB features and depth features to generate semantic and contextual features. We then compute the 4D correlation volume via computing the dot product of the fused features according to different location pairs.

\subsection{Regressing 3D Transformation Field}
\label{subsec:se3motiton}

With a dense matching field as the index, we can extract a 3D correlation map from the 4D correlation volume.

In SCFlow~\cite{hai2023scflow}, the correlation map is used to predict the optical flow $f(u,v)$, which maps the pixel position $\mathbf{x}_{1}=(u,v)$ in $I_{1}$ to the corresponding position in $I_{2}$: $\mathbf{x}_{2}=(u,v)+f(u,v)$. However, this 2D flow formulation only contains 2D information, which is less effective in capturing 3D motions.

To address this, we propose to predict 3D scene flow~\cite{teed2021raft3d} as an intermediate representation for the network.
Given $I_1$ and $I_2$ and their corresponding point clouds $C_1$ and $C_2$, we formulate scene flow as a dense 3D transformation field $\mathbf{T}\in SE(3)^{H\times W}$ to represent the 3D motion between the two point clouds in 3D space.

Specifically, for the point $X_{i}$ in $C_1$ with index $i$ and its projected pixel coordinates $\mathbf{x}_i=(u_i,v_i,z_i)$, we can perform the transformation $\mathbf{T}_i\cdot X_i$ to obtain the corresponding point $X_i^{\prime}$ in $C_2$ and its projected pixel coordinates $\mathbf{x}_i^{\prime}=(u_i',v_i',z_i')$. The scene flow vector is defined as the difference $\mathbf{f} = \mathbf{x}_i^{\prime}-\mathbf{x}_i$, where the first two components represent the standard 2D optical flow and the last component describes the depth difference between the two points.

We use GRU~\cite{GRU} as the recurrent model for the intermediate scene flow predictor. In the $k^{th}$ iteration, the hidden state update can be expressed as:
\begin{equation}
h_{k}=\mathrm{GRU}(\mathbf{L}_{\mathbf{C}}(F_{k-1}),\mathbf{T}_{k-1},h_{k-1};\Theta)
\end{equation}
where $\mathbf{L}_{\mathbf{C}}(F_{k-1})$ represents the correlation look-up operation guided by the global flow $F_{k-1}$ from the 
$(k-1)^{th}$ iteration, $\mathbf{T}_{k-1}$ is the dense 3D transformation field from the previous iteration, $\Theta$ denotes the GRU network parameters, and $h_{k}$ is the hidden state feature
which will be fed into a dense SE3 layer~\cite{teed2021raft3d} to update the SE3 motion fields and generate a new transformation field $\mathbf{T}_{k}^{\prime}$. 

\subsection{Regressing Global Object Pose}

Our goal is to predict the pose residual of the single object in $I_1$ and $I_2$. Considering the rigidity of objects in 6D pose estimation scenarios, all pixel-level 3D motions in $\mathbf{T}_{k}^{\prime}$ theoretically describe the same rigid object and should be the same. However, noise is inevitable in predicting the dense transformation field.
Therefore, we designed a small network to predict the global pose residual using the dense transformation field as input, as shown in \cref{fig:scene_flow}. First, we represent the dense transformation field $\mathbf{T}_{k}^{\prime}$ as a $4 \times 4$ transformation matrix, and then we encode it using a three-layer 2D convolutional network. Finally, we use two fully connected layers to output the global pose residual $\Delta\mathbf{P}_{k}$, which is represented as a 9-dim vector including 6-min for the rotation~\cite{6Drotation} and 3-dim for the normalized translation~\cite{li2018deepim}.

In the $k^{th}$ iteration, the pose residual $\Delta\mathbf{P}_{k}$ will be used for the pose update: $\mathbf{P}_{k}=\mathbf{P}_{k-1}\otimes\Delta\mathbf{P}_k$, and the updated pose $\mathbf{P}_k$ will be used to compute the pose-induced flow based on the object mesh, which imposes shape constraints for the lookup operation in the next iteration~\cite{hai2023scflow}. At the same time, we will generate a new dense transformation field $\mathbf{T}_k$ based on the updated pose $\mathbf{P}_{k}$. Specifically, we calculate the pose residual between the initial pose $\mathbf{P}_0=[\mathbf{R}_0|\mathbf{t}_0]$ and $\mathbf{P}_k=[\mathbf{R}_k|\mathbf{t}_k]$ as follows:
\begin{equation}
\Delta \mathbf{R}_k=\mathbf{R}_k\cdot\mathbf{R}_0,\quad\Delta \mathbf{t}_k=\mathbf{t}_k-\Delta\mathbf{R}_k\cdot\mathbf{t}_0
\end{equation}
Then, we update the 3D motion field $\mathbf{T}_k\in SE(3)^{H\times W}$, where the motion at every location will be updated as the same new pose residual. This new motion field will be the input for the next iteration.

\begin{table*}[htbp]
  \centering
  \resizebox{0.8\linewidth}{!}{
    \begin{tabular}{lcccccccc}
    \toprule
     Methods & LM-O & T-LESS & TUD-L & IC-BIN & ITODD & HB & YCB-V & Avg. \\
    \midrule
    MegaPose~\cite{labbe2022megapose} & 62.0  & 48.5  & 84.6  & 46.2  & 46.0  & 72.5  & 76.4  & 62.3  \\
    MegaPose + {\bf Ours} & {\bf 62.4}  & \textbf{54.9}  & \textbf{88.3}  & \textbf{48.4}  & \textbf{49.5}& \textbf{76.4}  & \textbf{79.0}  & \textbf{65.6}  \\
    \midrule
    FoundPose~\cite{ornek2023foundpose} & 61.0  & 57.0  & 69.4  & 47.9  & 40.7  & 72.3  & 69.0  & 59.6  \\
    FoundPose + {\bf Ours} & \textbf{69.6}  & \textbf{58.6}  & \textbf{88.1}  & \textbf{54.6}& \textbf{48.4}  & \textbf{76.2}  & \textbf{84.9}& \textbf{68.6}  \\
    \midrule
    GenFlow~\cite{moon2024genflow} & 63.5  & 52.1  & 86.2  & 53.4  & 55.4  & 77.9  & 83.3  & 67.4  \\
    GenFlow + {\bf Ours} & \textbf{68.7}  & \textbf{53.7}  &\textbf{88.0}  & \textbf{55.4}  &\textbf{56.4}& \textbf{79.2  }& \textbf{85.5}& \textbf{69.6} \\
    \midrule
    Gigapose~\cite{nguyen2024gigapose} & 67.8  & 55.6  & 81.1  & 56.3  & 57.5  & 79.1  & 82.5  & 68.6  \\
    Gigapose + {\bf Ours} & \textbf{71.0} & \textbf{58.6} & \textbf{82.0} & \textbf{57.7} & \textbf{58.3} & \textbf{80.1} & \textbf{84.4} & \textbf{70.3} \\
    \midrule
    SAM6D~\cite{lin2024sam6d} & 69.9  & 51.5  & 90.4  & 58.8  & 60.2  & 77.6  & 84.5  & 70.4  \\
    SAM6D + {\bf Ours} & \textbf{72.1} & \textbf{52.3} & \textbf{90.8} & \textbf{60.0} & \textbf{60.5} & \textbf{77.7} & \textbf{85.1} & \textbf{71.2} \\
    \midrule
    FoundationPose~\cite{wen2024foundationpose} & 75.6  & 64.6  & 92.3  & 50.8  & 58.0  & 83.5  & 88.9  & 73.4  \\
    FoundationPose + {\bf Ours} & \textbf{80.1} & \textbf{65.0} & \textbf{95.5} & \textbf{51.7} & \textbf{58.3} & \textbf{84.4} & \textbf{91.4} & \textbf{75.2} \\
    \bottomrule
    \end{tabular}}
    \caption{\textbf{Comparison with the state of the art.} All the baselines have their own refinement strategy, and we report their final result published on BOP leaderboard. After using our method as a post-processing (``+ Ours''), the result consistently improves across all baselines on all datasets.
    }
  \label{tab:compare_with_stoa}%
\end{table*}%

\subsection{Implementation Details}
\label{subsec:details}

For the input RGB and depth images $I_1$ and $D_2$, we first crop the target out based on the initial pose, and then resize it to a fixed resolution of 256$\times$256.

Given the ground truth pose, we can compute the ground truth optical flow based on the object mesh. We use the exponentially weighted strategy~\cite{teed2020raft, hai2023scflow} to calculate the loss in each iteration for both flow loss and pose loss:
\begin{equation}
\mathcal{L}=\sum_{k=1}^N\gamma^{N-k}(\mathcal{L}_{pose}^k+\alpha\mathcal{L}_{flow}^k)
\end{equation}
where the weight $\gamma = 0.8$, $N = 8$ and $\alpha = 0.1$. $\mathcal{L}_{flow}$ represents the L1 loss between the first two dimensions of the predicted scene flow and the ground truth optical flow. $\mathcal{L}_{pose}$ refers to pose loss, which is computed as the L1 distance between the 3D point clouds transformed by the predicted pose and the ground truth pose, respectively~\cite{wang2021gdr, hai2023scflow}.

\section{Experiments}
\label{sec:experiments}

We evaluate SCFlow2 on seven challenging BOP benchmark datasets in this section. \Cref{subsec:set} presents our experimental setup, including the datasets used for training and evaluation, training details and the evaluation metrics. In \Cref{subsec:cpsota}, we compare our approach with current state-of-the-art pose methods. \Cref{subsec:ablation} provides a detailed ablation study of the proposed framework.

\subsection{Experimental Settings}
\label{subsec:set}

\noindent \textbf{Dataset}. We train SCFlow2 on a combination of datasets including ShapeNet-Objects~\cite{chang2015shapenet}, Google-Scanned-Objects~\cite{downs2022gso}, and Objaverse~\cite{deitke2023objaverse}. 
All these datasets only contain object meshes, in a total of about 90K objects. We reuse the synthetic images from~\cite{labbe2022megapose} and~\cite{wen2024foundationpose} which are rendered based on those meshes, with about 3M images in total.


We evaluate on seven datasets from the BOP benchmark~\cite{hodan2018bop}, including LM-O~\cite{lmo}, T-LESS~\cite{hodan2017tless}, TUD-L~\cite{hodan2018bop}, IC-BIN~\cite{doumanoglou2016icbin}, ITODD~\cite{drost2017itodd}, HB~\cite{kaskman2019hb}, and YCB-V~\cite{xiang2017posecnnycbv}. These datasets include 108 objects having different sizes and different texture and symmetry properties. Most of them are captured in challenging setups, including multiple object instances, significant clutter, and severed occlusions. These objects are not included in the training dataset and are novel to the models.

\noindent \textbf{Training details}. We train SCFlow2 using the AdamW optimizer~\cite{ADAMW} with a batch size of 16, and use a cosine annealing learning rate scheduler. The model is trained for 200k iterations with an initial learning rate of 1e-4. During training, we generate pseudo initial poses by adding random noise to the ground-truth pose. We add rotation jitter sampled from a normal distribution with a standard deviation of 15 degrees in each 3D axis, and add translation jitter sampled with deviations of 15, 15, and 50 millimeters along xyz axes, respectively.

\noindent \textbf{Evaluation metrics} We report the result in the common BOP metric, which is the mean Average Recall (AR) of three different metrics, including Visible Surface Discrepancy (VSD), Maximum Symmetry-aware Surface Distance (MSSD), and Maximum Symmetry-aware Projection Distance (MSPD). Readers can refer to \cite{hodan2018bop} for further details about these metrics.

\subsection{Comparison with the State of the Art}
\label{subsec:cpsota}

To explore the plug-and-play nature of our refinement method, we apply it to the final results of Megapose~\cite{labbe2022megapose}, FoundPose~\cite{ornek2023foundpose}, GenFlow~\cite{moon2024genflow}, 
GigaPose~\cite{nguyen2024gigapose},
SAM6D~\cite{lin2024sam6d},
and FoundationPose~\cite{wen2024foundationpose}, as shown in Table~\ref{tab:compare_with_stoa}. Although these baselines have a refinement strategy themselves, plugging our refinement method into the end of their framework further improves the results, consistently on all datasets. Most baseline results are downloaded from the BOP leaderboards.


\subsection{Ablation Study}
\label{subsec:ablation}

\begin{figure*}[tp]
    \centering
    \setlength\tabcolsep{1pt}
    \begin{tabular}{ccccccc}
    \includegraphics[width=0.16\linewidth]{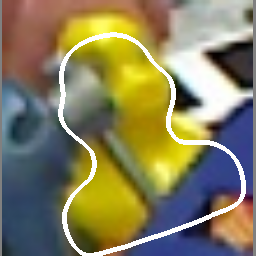} &
    \includegraphics[width=0.16\linewidth]{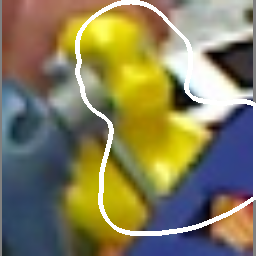} &
    \includegraphics[width=0.16\linewidth]
    {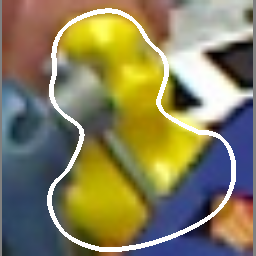} &
    \includegraphics[width=0.16\linewidth]{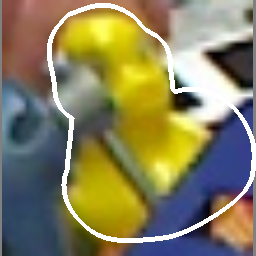} &
    \includegraphics[width=0.16\linewidth]{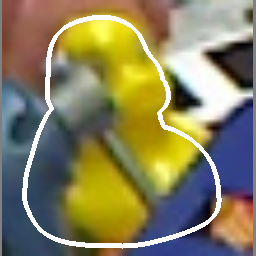} &
    \includegraphics[width=0.16\linewidth]{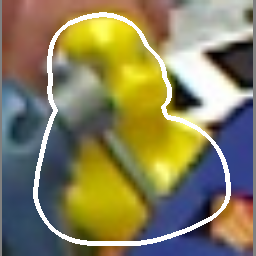} \\
        \includegraphics[width=0.16\linewidth]{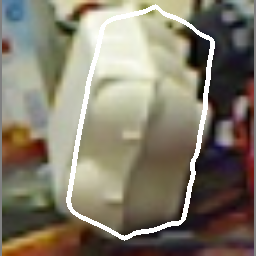} &
    \includegraphics[width=0.16\linewidth]{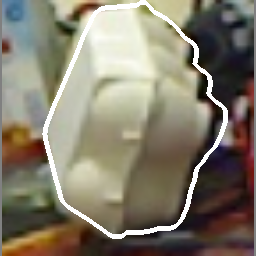} &
    \includegraphics[width=0.16\linewidth]
    {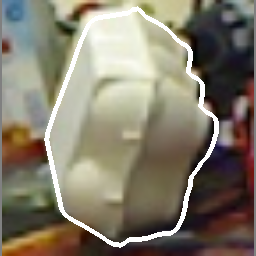} &
    \includegraphics[width=0.16\linewidth]{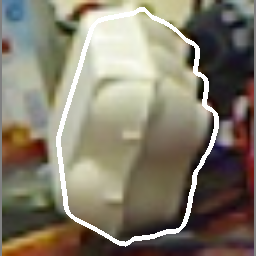} &
    \includegraphics[width=0.16\linewidth]{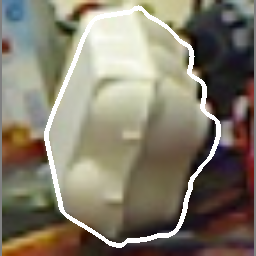} &
    \includegraphics[width=0.16\linewidth]{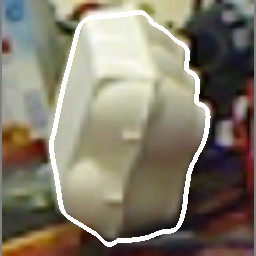} \\
        \includegraphics[width=0.16\linewidth]{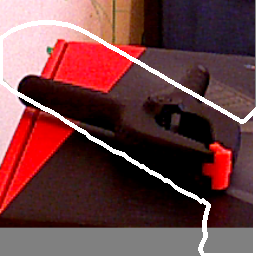} &
    \includegraphics[width=0.16\linewidth]{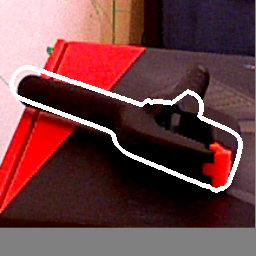} &
    \includegraphics[width=0.16\linewidth]
    {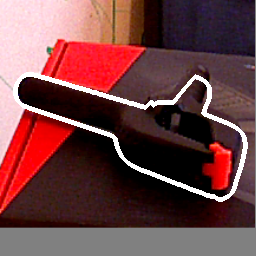} &
    \includegraphics[width=0.16\linewidth]{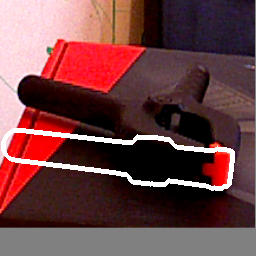} &
    \includegraphics[width=0.16\linewidth]{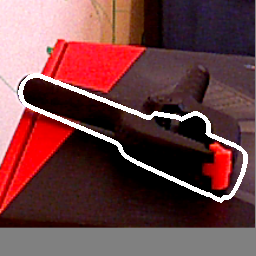} &
    \includegraphics[width=0.16\linewidth]{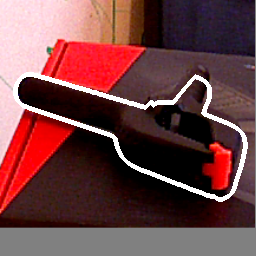} \\
        \includegraphics[width=0.16\linewidth]{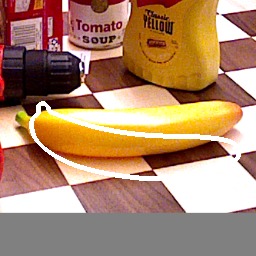} &
    \includegraphics[width=0.16\linewidth]{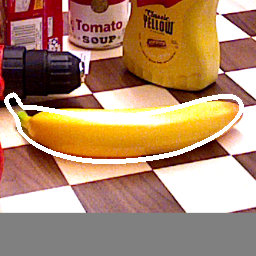} &
    \includegraphics[width=0.16\linewidth]
    {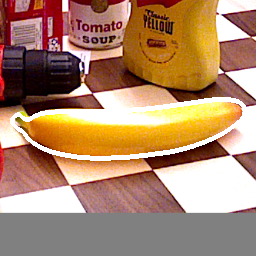} &
    \includegraphics[width=0.16\linewidth]{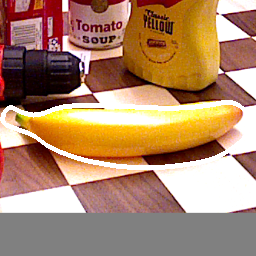} &
    \includegraphics[width=0.16\linewidth]{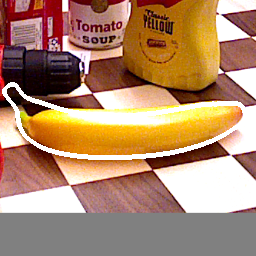} &
    \includegraphics[width=0.16\linewidth]{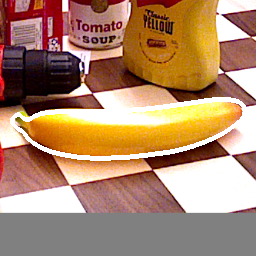} \\
        \includegraphics[width=0.16\linewidth]{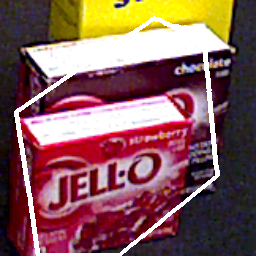} &
    \includegraphics[width=0.16\linewidth]{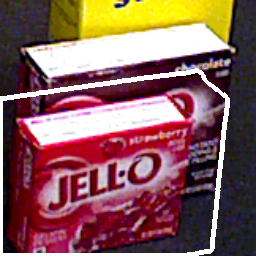} &
    \includegraphics[width=0.16\linewidth]
    {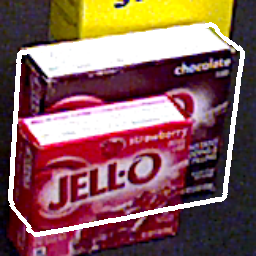} &
    \includegraphics[width=0.16\linewidth]{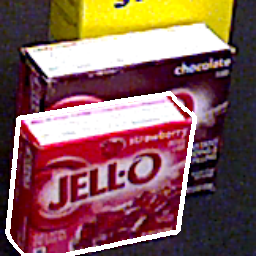} &
    \includegraphics[width=0.16\linewidth]{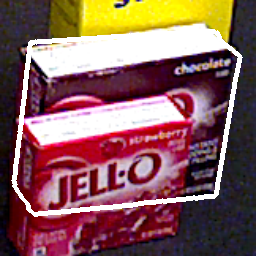} &
    \includegraphics[width=0.16\linewidth]{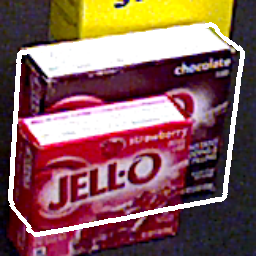} \\

        \includegraphics[width=0.16\linewidth]{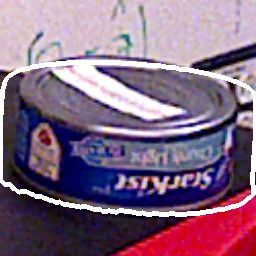} &
    \includegraphics[width=0.16\linewidth]{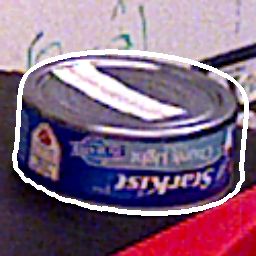} &
    \includegraphics[width=0.16\linewidth]
    {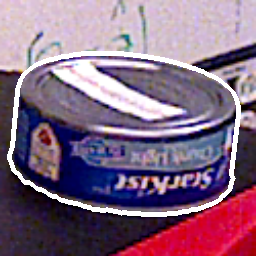} &
    \includegraphics[width=0.16\linewidth]{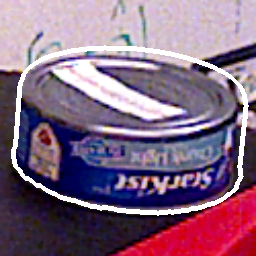} &
    \includegraphics[width=0.16\linewidth]{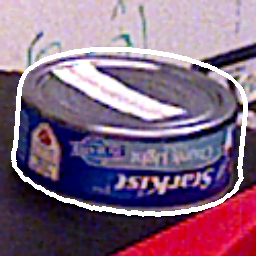} &
    \includegraphics[width=0.16\linewidth]{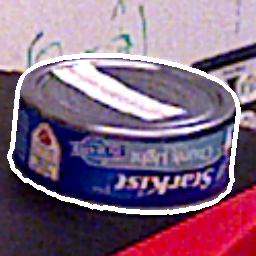} \\
    
    {\small GFlow (init)} & {\small GFlow (final)} & 
    \multicolumn{1}{c}{{\cellcolor{green!10} {\small GFlow (init) + {\bf Ours}}}} &
    {\small FPose (init)} & {\small FPose (final)} & 
    \multicolumn{1}{c}{{\cellcolor{green!10} {\small FPose (init) + {\bf Ours}}}} \\
    \end{tabular}
    \caption{{\bf Qualitative results.} Given the same pose initialization as that in GenFlow (``GFlow'') and FoundPose (``FPose''), denoted as ``GFlow (init)'' and ``FPose (init)'' respectively, our refinement method (``+ Ours'') produces considerably more accurate results compared to the refinement approaches in their original methods (note how our reprojected 3D mesh aligns better with the object contours).
    }
    \label{fig:final_qualitative}
\end{figure*}

\begin{figure}
    \centering
    \setlength\tabcolsep{1pt}
    \begin{tabular}{cccc}

    \includegraphics[width=0.48\linewidth]{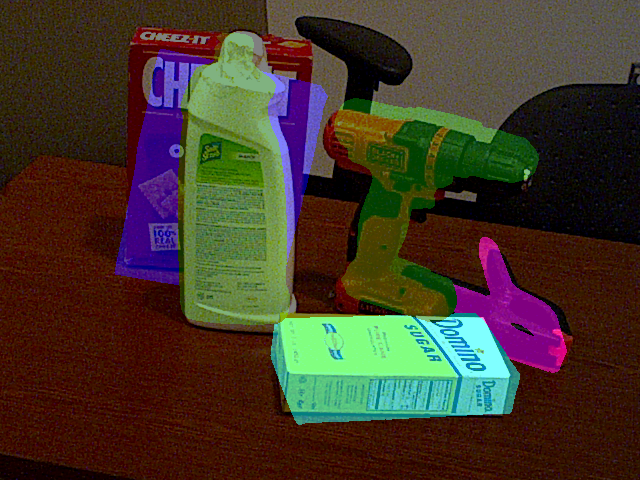} &
    \includegraphics[width=0.48\linewidth]{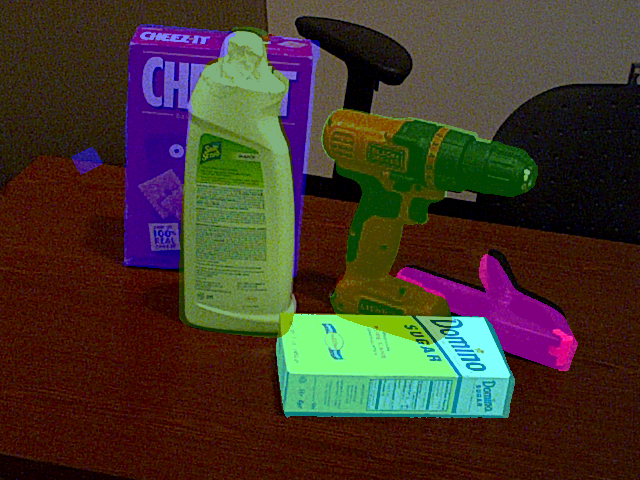} \\

    \includegraphics[width=0.48\linewidth]{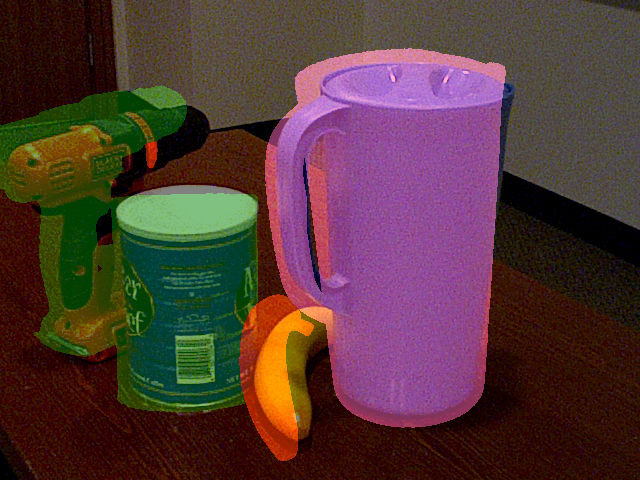} &
    \includegraphics[width=0.48\linewidth]{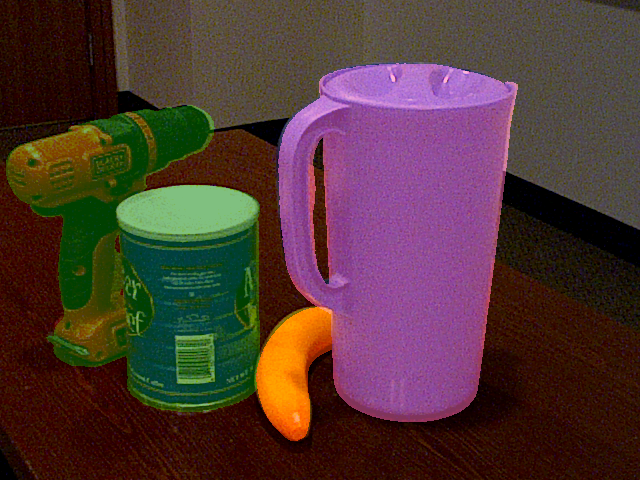} \\
    

    
    {\small Initialization} & {\small + {\bf Ours}}
    \end{tabular}

    \caption{{\bf More qualitative results.}
    We show two examples on YCB-V. Although the initialization has large pose error, our refinement method produces accurate results that align well with the target, without retraining on these novel objects.
    }
    \label{fig:add_qualitative}
\end{figure}

\noindent {\bf Effectiveness as a plug-and-play object pose refiner.}
We evaluate SCFlow2 in more detail in the context of three different state-of-the-art methods, FoundPose~\cite{ornek2023foundpose}, GenFlow~\cite{moon2024genflow}, and FoundationPose~\cite{wen2024foundationpose}. Table~\ref{tab:compare_refine_ablation} summaries the results.
All of them are based on an initialization-refinement architecture. We denote their initialization results as ``init'' in the table, and replace their refinement methods with ours in each method (denoted as ``init + ours''). Furthermore, we also report the results of applying our refinement to their final results directly, denoted as ``final + ours''. Note that, FoundationPose outputs hundreds of pose hypotheses in the initialization stage, and there is no well-defined initialization result for it. In this case, we replace its original refinement method with SCFlow2 in its framework, and keep all other components untouched, including post-processing procedures such as pose scoring and pose selection, which is the reported result of ``FoundationPose (init) + Ours'' in the table. Overall, our method consistently outperforms their original refinement strategies, and can also improve their final results significantly, in a plug-and-play manner. We show some qualitative results in Fig.~\ref{fig:final_qualitative} and~\ref{fig:add_qualitative}.

\begin{table}[t]
  \centering
  \resizebox{1.0\linewidth}{!}{
    \begin{tabular}{lccc}
    \toprule
    Methods & Depth & LM-O & YCB-V \\
    \midrule
    FoundPose (init) & \xmark & 39.6  & 45.2  \\
    FoundPose (final) & \cmark & 61.0  & 69.0  \\
    FoundPose (init) + {\bf Ours} & \cmark & \underline{61.3} & \underline{81.8} \\
    FoundPose (final) + {\bf Ours} & \cmark & \textbf{69.6} & \textbf{84.9} \\
    \midrule
    GenFlow (init) & \xmark & 25.0  & 27.7  \\
    GenFlow (final) & \cmark & 63.5  & 83.3  \\
    GenFlow (init) + {\bf Ours}  & \cmark & \underline{64.4} & \underline{83.6} \\
    GenFlow (final) + {\bf Ours} & \cmark & \textbf{68.7} & \textbf{85.5} \\
    \midrule
    FoundationPose (init) & - & -  & -  \\
    FoundationPose (final) & \cmark & 75.6  & 88.9  \\
    FoundationPose (init) + {\bf Ours} & \cmark & \underline{79.7} & \underline{90.4} \\
    FoundationPose (final) + {\bf Ours} & \cmark & \textbf{80.1} & \textbf{91.4} \\
    \bottomrule
    \end{tabular}
    }
    \caption{\textbf{Effectiveness as a plug-and-play object pose refiner.}
    All the baselines are based on an initialization-refinement architecture. Our method consistently outperforms their original refinement strategies, and can also improve their final results significantly, in a plug-and-play manner.
    }
  \label{tab:compare_refine_ablation}
\end{table}%

\begin{table}[t]
  \centering
  \scalebox{0.88}{
    \begin{tabular}{cccccccc}
    \toprule
     & L3 & L5 & L10 & L20 & L30 & L40 & L50 \\
    \midrule
     Initialization & 92.7 & 83.8 & 64.5 & 35 & 17.6 & 7.8 & 3.5 \\
     + SCFlow2 & 91.6 & 91.5 & 91.5 & 90.8 & 88.6 & 83.2 & 70.7 \\
    \bottomrule
    \end{tabular}%
    }
  \caption{{\bf Impact of pose initialization.} We add randomly generated rotation and translation noise to GT pose as the initialization, and then use SCFlow2 to refine the pose. L3 to L50 denote different noise levels. SCFlow2 is robust to pose errors in the initialization and works well under noise level 30.
  } 
  \label{tab:init_pose_quality}%
\end{table}%

\begin{table}[]
  \centering
    \begin{tabular}{lc|cccc}
    \toprule
    &&\multicolumn{2}{c}{Seen} & \multicolumn{2}{c}{Unseen} \\
    Datasets & Init. & V1 & V1+ & V1++ & \textbf{V2} \\
    \midrule
    LM-O & 61.0  & 62.9  & \underline{69.3}  & 68.9  & \textbf{69.6} \\
    T-LESS & 57.0  & \underline{57.9}  & 54.9  & 55.3  & \textbf{58.6} \\
    TUD-L & 69.4  & 75.9  & \textbf{88.8} & 86.6  & \underline{88.1}  \\
    IC-BIN & 47.9  & 47.9  & \underline{54.1}  & 53.7  & \textbf{54.6} \\
    ITODD & 40.7  & 34.2  & \textbf{49.1} & 48.1  & \underline{48.4}  \\
    HB & 72.3  & 74.5  & \textbf{76.5} & 76.1  & \underline{76.2}  \\
    YCB-V & 69.0  & 80.7  & \textbf{85.5} & 84.4  & \underline{84.9}  \\
    \midrule
    Avg. & 59.6  & 62.0  & \underline{68.3}  & 67.6  & \textbf{68.6} \\
    \bottomrule
    \end{tabular}%
  \caption{\textbf{Advantages over SCFlow.} We use the same initialization based on results from FoundPose (``Init.''), and report results of SCFlow (``V1'') and SCFlow2 (``V2'').
  SCFlow2 outperforms all ``V1'' versions without any retraining.
  }
  \label{tab:compare_scflow1}
\end{table}%

\noindent {\bf Impact of pose initialization.}
We add random rotation and translation noise to GT pose as the initialization, and report the results on the YCB-V dataset, as shown in Table~\ref{tab:init_pose_quality}. L3 to L50 denote different noise levels (rotation error from 3 to 50 degrees, translation error from 3 to 50 mm). SCFlow2 works well under noise level 30.



\begin{figure}[t]
\centering
    \includegraphics[width=0.48\linewidth]{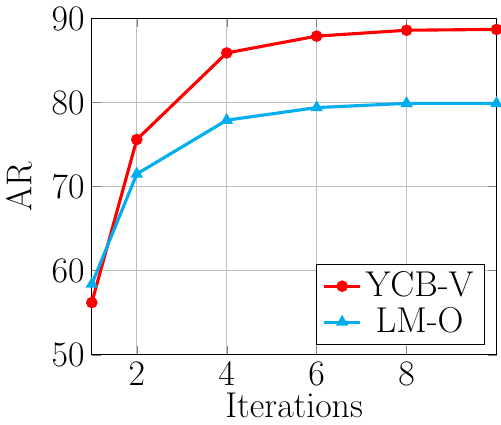}
        \hspace{0.2em}
    \includegraphics[width=0.48\linewidth]{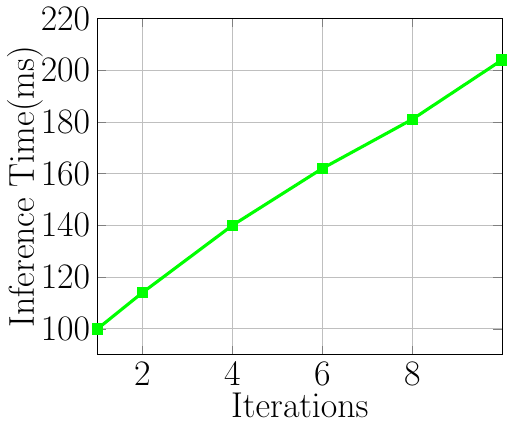} 
  \caption{{\bf Evaluation of performance with different iterations.} Given the same pose initialization with noise level 30 as in Table~\ref{tab:init_pose_quality}, more recurrent iterations produce more accurate results while spending more time in inference. Nevertheless, our approach yields meaningful outcomes after just 2 iterations, and the performance reaches saturation after 6 iterations.
  }
  \label{fig:diff_iter}
\end{figure}

\noindent {\bf Evaluation of performance with different iterations.}
We conduct ablation studies on the trade off between the accuracy/efficiency and the iteration numbers of our method, as shown in~\cref{fig:diff_iter}. Our method produces meaningful pose refinement results after only 2 iterations and reaches saturation after 6 iterations. In our experiments, we use 8 iterations as the default setting in both training and testing.


\noindent \textbf{Advantages over SCFlow.}
We compare SCFlow2 and its previous version SCFlow systematically. We summarize the results in Table~\ref{tab:compare_scflow1}. We use the same initialization based on results from FoundPose, and refine it with SCFlow (``V1'') and SCFlow2 (``V2''), respectively. We evaluate three different versions of SCFlow, where ``V1'' denotes the original version based on retraining on target objects with RGB images only, ``V1+'' denotes the version using RANSAC Kabsch to consume additional depth images. Both ``V1'' and ``V1+'' need retraining on target objects. We also evaluate a version ``V1++'' which does not rely on retraining for novel objects, but uses the same large-scale datasets as ``V2'' for training on RGB images and then refine the pose results using RANSAC Kabsch with additional depths. We can see that, without retraining, ``V1++'' produces worse results compared with ``V1+''. On the other hand, SCFlow2 even outperforms ``V1+'' without any retraining.

\noindent {\bf Evaluation with different settings.}
We use the initialization results of FoundPose as our initialization.
Table~\ref{tab:ablation} shows the result, where ``Shape'' denotes the shape-constraint regularization during the recurrent optimization, and ``Scene'' is the 3D scene flow representation. We use the intermediate flow for the look-up without relying on the target's 3D mesh when ``Shape'' is disabled, and we replace the scene flow representation with a plain MLP pose regressor when ``Scene'' is disabled. The performance becomes worse without the shape prior embedding, and deteriorates significantly without the 3D scene flow formulation.

\begin{table}[]
  \centering
    \begin{tabular}{cccc}
    \toprule
    Shape & Scene & LM-O & YCB-V \\
    \midrule
    \cmark & \xmark & 52.6  & 67.6  \\
    \xmark & \cmark & \underline{60.0}  & \underline{80.4}  \\
    \cmark & \cmark & \textbf{61.3} & \textbf{81.8} \\
    \bottomrule
    \end{tabular}%
    
  \caption{\textbf{Evaluation with different settings.}
  ``Shape'' denotes the shape-constraint regularization during the optimization, and ``Scene'' is the 3D scene flow formulation. The performance deteriorates significantly without the 3D scene flow formulation.
  }
  \label{tab:ablation}%
\end{table}%
\noindent {\bf Timing analysis.}
The training of SCFlow2 needs about 108 hours on a workstation with 2 NVIDIA RTX-3090 GPU. SCFlow2 considers the pose initialization as the only hypothesis. During inference, SCFlow2 takes about 0.18 seconds to refine a pose on an RTX-3090 GPU with a typical image from BOP dataset. Taken for comparison, the multi-hypothesis based method MegaPose needs about 1.85 seconds for a refinement on the same workstation, which is 10+ times slower than SCFlow2.

\begin{figure}[]
\centering
    \includegraphics[width=0.48\linewidth]{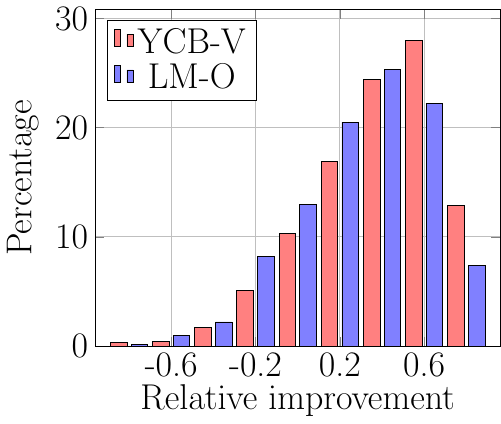} 
  \caption{{\bf Limitation discussion.} We show the histogram of samples along with the relative accuracy improvement after pose refinement. In the majority of cases, our method improves the initial poses significantly. However, it still happens that the refinement makes the results worse in some extreme cases.
  }
  \label{fig:limitation}
\end{figure}

\noindent {\bf Limitation discussion.}
In general, our refinement method can improve the accuracy of initial pose significantly. While occasionally, it produces worse results than the initial pose. Fig.~\ref{fig:limitation} shows the histogram of samples along with the relative accuracy improvement after pose refinement on YCB-V and LM-O with the initial pose of FoundPose as our initialization. In the majority of cases, our method improves the initial poses significantly. However, it still happens that the refinement makes the results worse in some extreme cases where the initial poses are already very wrong caused by severe occlusion or extreme light conditions. We will take this as one of our future work.

\section{Conclusion}
\label{sec:conclusion}

We have introduced SCFlow2, a plug-and-play refinement framework for 6D object pose estimation. We extended the RGB-based method SCFlow by formulating additional depth as a regularization in the iteration with an intermediate 3D scene flow representation for RGBD frames, and combined it with the 3D shape prior information of the target, producing a geometry-guided 6D object pose refiner end-to-end trainable and generalizable to novel objects. Experiments on BOP datasets with novel objects show that our method achieves state-of-the-art accuracy, and is one of the most efficient refinement methods.

\noindent {\bf Acknowledgments.}
This work was supported in part by the Youth Innovation Team of Shaanxi Universities, the ``Scientist + Engineer'' team of the Qin Chuang Yuan in Shaanxi Province, the Xi'an Science and Technology Program, the ``Leading the Charge'' initiative for the industrialization of core technologies in key industrial chains in Shaanxi Province, the National Nature Science Foundation of China under Grant 62371359, the Key Research and Development Program of Shaanxi, the Key Research Program of the Chinese Academy of Sciences.

{
    \small
    \bibliographystyle{ieeenat_fullname}
    \bibliography{main}
}


\end{document}